# Super-resolution Reconstruction of SAR Image based on Non-Local Means Denoising Combined with BP Neural Network

Zeling Wu[1,2*]; Haoxiang Wang[3,4]
[1]East China University of Science and Technology, China;
[2]Lüebeck University of Applied Science, Germany, Lüebeck, Germany
[3]Department of ECE, Cornell University, NY, USA
[4] GoPerception Laboratory, NY, USA
*zeling.wu@stud.fh-luebeck.de

*Abstract*—In this article, we propose a super-resolution method to resolve the problem of image low spatial because of the limitation of imaging devices. We make use of the strong non-linearity mapped ability of the back-propagation neural networks(BPNN). Training sample images are got by undersampled method. The elements chose as the inputs of the BPNN are pixels referred to Non-local means(NL-Means). Making use of the self-similarity of the images, those inputs are the pixels which are pixels gained from modified NL-means which is specific for super-resolution. Besides, small change on core function of NL-means has been applied in the method we use in this article so that we can have a clearer edge in the shrunk image. Experimental results gained from the Peak Signal to Noise Ratio(PSNR) and the Equivalent Number of Look(ENL), indicate that adding the similar pixels as inputs will increase the results than not taking them into consideration.

## I. INTRODUCTION

Remotely sensed images are images of the target on the ground which contains the features taken from different angles and heights. There are many ways to acquire remotely sensed images, one of which is through Synthetic Aperture Radar(SAR). SAR is a system which can observe the landscape and typically mounted on flying platform like airplanes, satellites and spacecrafts. Besides, SAR provides observation regardless of weather and seasons. Because of these advantages, SAR plays an important role in disaster monitoring, environmental monitoring, resource exploration etc. But, because of the limitation of imaging devices, the low spatial resolution is always a problem. Besides, SAR images are suffering from the severe influence of speckle noise. Speckle noise is multiplicative and nonwhite process which means that many super resolution reconstruction algorithms won't be a proper choice.

The methods used by the scientists can be ususally divided into two types: one is the super-resolution reconstruction based on multiple images and the other one is the super-resolution reconstruction based on a single image. It is indeed that super-resolution reconstruction based on multiple images obtains higher accuracy and better performance in most of the conditions. However, in reality, multiple low resolution images with subpixel-shifted are not easily gained. So this article mainly focus on super-resolution based on single image.

Neural network(NN) is an abstract model of our brain which constructs of calculating units connecting with one another. Although modern neural network project work with thousands of neural units, neural network is still not compatible of our brain. Neural network is widely used in many fields of SAR, for example, classification and recognition of the target picture. Neural network provides high performance on nonlinear statistical modeling, so it may work well as an alternative on reconstructing SAR.

Back Propagation(BP) Algorithm is one of the most popular NN algorithms which is capable of finding correct combination of weights for the NN to deal with complex problems. With its strong nonlinearity mapping ability, it is very helpful in finding the hidden relationships between the inputs and the outputs. In this paper, BP Algorithm has played a very important role in proving the relationships between the pixels between the largest similarity between the target pixel and the result pixels.

The remaining of the paper is organized as follows. In the section 2, we will introduce the previous research briefly. In the third section, we will look into the method we have proposed. In the fourth section, we will demonstrate the algorithm more by showing the experiments' results. In the last section, we will draw a conclusion to this thesis.

## II. LITERATURE REVIEW

### A. Super-Resolution based on Excluded Aliased Frequency Domain

The earliest work reconstructing high-resolution image through frequency domain traces back to Tsai and Huang, in 1984[1]. Assuming that the band width of the signal of the original scene is limited, a formula which is able to reconstruct high-resolution image through a series of low-resolution images is gained. This equation which is based on the shift and aliasing properties of the Continuous and Discrete Fourier Transforms relates the Discrete Fourier Transforms of those low-resolution images and the Continuous Fourier Transforms of high-resolution images. The result of this equation is the Continuous Fourier Transforms operator of the high-resolution image. Finally, gaining the inverse Fourier transform of it will be the high resolution image. To implement this algorithm, we have to ensure that different low resolution images must have different sub-pixel shift information so that every image can provide a linear independent equation. Although this method can be easily calculated, it fails to consider the influence of sensor blurring and noise. So the direct use of this method is limited.

*B. Image Super-Resolution based on Sparse Representation of Image*

Kim proposed Weighted least squares[2] to reconstruct low resolution images based on the existing frequency algorithm after carefully analyze the noise. Although this algorithm is robust and effective, they fail to discuss question like non-global shift and movement compensated estimation. So, Su and Kim[3] made a further improvement by proposing the algorithm on frequency that combines non-global motion model. To be specific, the observed low resolution image will be segmented into small pieces, then we will apply motion estimation on these pieces separately. Finally, we will reconstruct small pieces and combine them into the high resolution image we needed. However, this algorithm still uses global motion model when reconstructing each segment, in other words still fails to solve the problem of non-global shift.

*C. Bicubic Interpolation Enlargement Algorithm*

There are three main interpolation methods: Nearest-Neighbor interpolation, Bilinear interpolation and Bicubic interpolation. Bicubic interpolation is the slowest algorithm, however is the one which provides the best effect. It provides smoother edge than the other two methods. It implements three interpolations using 16 points surrounded with the target pixel. It not only takes the gray scale into consideration, but also the changing rate between every pixel. To implement Bicubic Interpolation. We have to firstly decide our basis function. Often a cubic function is chosen. Then value and derivation of the basis function at 0 and 1 can be easily gained. If we estimate that curve between -1 to 2 is almost a line, we can express the derivation with two pixels. Assuming it is a straight line, we can understand this better. We can get the value of target point with 4 pixels in x-plane. Then, we may use 4 pixels in y-plane to get the pixel in x-plane in the same way. Finally, the target value in xy-plane is gained with the value of 16 surrounding pixels .

*D. Image Super-Resolution based on Sparse Representation of Image*

Sparse Representation is a technique which has long been used in image compress even since a thesis of historical significance published in 1996[4] has shown that human visual system can capture the key information of a image with minimum visual neurons. In other word, after some kinds of transform, image can be expressed in a way that only a few elements are not zero, but almost all elements are zero or nearly zeros. And soon after this thesis, they proposed a completed algorithm based on sparse coding. An image can be expressed as the product of a redundant dictionary and a sparse vector. We call every element in the dictionary a atom. The purpose of the algorithm is that the image should be sparse under the representation of the linear combination of those atoms. This dictionary can be gained from two methods. One is based on analysis. these dictionaries have very good structure and many algorithm has been proposed to accelerate the speed to gain these dictionaries. However, they perform poorly facing different types of data. The other one is based on learning. Those dictionaries are gained from some samples. Although those dictionaries are well-adaptive, they are heavily redundant and slow.

*E. Image Super-Resolution based on Iterative Back Projection(IBP)*

Iterative Back Projection is proposed in 1989 by IRANI and PELEG[5]. This algorithm gains high resolution image by projecting HR image to LR image and feeding back the difference between simulated LR images and the observed LR images iteratively. Iterative Back Projection firstly uses an initial estimation of the output image as the current result and project it to the LR image. The difference between the project image and the real image is computed. The initial estimation is improved by "back-projecting " approach. By using error residual, an updated estimate of the original is gained. After every iteration, the approximation will become better. Assuming the noise is uncorrelated and has the same variance, this process will continue iteratively until the error function arrives at its minimum

*F. Image Super-Resolution based on Projection onto Convex Set(POCS)*

The convex set in this theory is referred to the constraints sets which consists of all the points which fulfill all the constraints. The base of this theory is that all the constraint sets are convex[6]. In this famous theory, every convex set denotes a solution to the constraints in desired image. In this case, the possible solution of reconstructing image must be inside the intersection. We can define the solution set of the problem to reconstruct the low-resolution image. We start from any point inside the space, keep searching until all the points of the intersection are found. Stark and Oskoui[7] proposed the very first high resolution image reconstruction algorithm based on Projection onto Convex Set. They defined a convex projection operator for every constraint sets. Then every initial estimation of the high-resolution image is projected every constraint sets. To achieve the best result, we can implement several times of iterative projections until we arrive at a satisfying high-resolution image. It is easy to use the POCS method to incorporate any kinds of constraints. However, it is slow and requires large amount of computation.

*G. Image Super-Resolution based on Maximum A Posteriori probability (MAP) estimate*

Talking about Maximum a Posteriori Probability (MAP) estimate, we can't neglect one algorithm close to it which is the method of Maximum Likelihood Estimation(MLE). MLE provides a method to estimate the model parameters through given observed data. In brief, assuming that we want to analyze a feature from a such great number of samples that we don't have the capability to check each of them, and this feature obeys normal distribution, but the mean value and variance of this feature remains unknown, we will still be able to analyze this feature by estimating these two values. We will take the features of only a part of the whole samples. Then we will come with the possibility represented by means and variance that we get the result from only gaining features of a small part. After that, the formula will go through a logarithm calculation so that we can carry out derivative more easily. And the means and variance have the greatest possibility to be true when the derivative reaches zero.

MAP has many in common with MLE. However, MLE is regardless of the model itself, and the MAP takes the model situation into consideration by multiplying the possibility that we arrive at this means and variance.

Schultz and Stevenson first proposed the combination of image reconstruction and MAP[8]. If we have express a high-resolution image by combination of some low-resolution image adding Gaussion noise, this expression is often morbid which may result in unstable solution. However, through applying MAP, we may solve this problem. In this algorithm, we will implement Markov Random Field(MRF) to calculate the possibilities mentioned above. This algorithm is rather slow because of the calculation it needed. We can improve it by implementing it iteratively or compute it in a parallel way.

III. METHODOLOGY

We firstly have to deal with the problem that SAR image which is full of multiplicative noise. We can apply our improved algorithm on the image, only after dealing with the noise. In our new algorithm, we will shrink image. In order to keep the edges of the image, a new core function will be proposed. Then the amplified image will be used as an input of the neural network.

A. Image Prepocessing

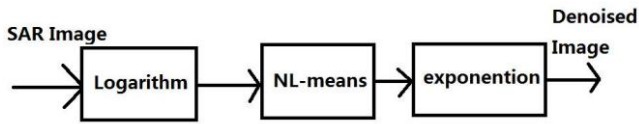

Fig1 the process of denoise

SAR image is full of multiplicative noise. However, NL-means is mainly used to remove Gaussion noise in the natural image. As the Gaussion noise is additive, we have to turn the calculation from Multiplication to addition. By using logarithm, we can apply NL-means for denoising the multiplicative noise in SAR image.

$$J = I + n*I \tag{1}$$

$$\log J = \log I + \log(n+1) \tag{2}$$

Assuming v(x) is image without noise, n(x) is the noise, and u(x) is the image we get.

$$u(v) = v(x)*n(x)$$

$$\log u(v) = \log v(x) + \log n(x)$$

And in the end, we will use Exponentiation to recover the image.

B. Non-local means (NL-means)

NL -means is a new denoise technique aroused these years. It makes use of redundancy of the image to maintain the main features of the image while denoising. Like "local mean" denoise, it calculates the average of the surrounding pixels, however, this algorithm also consider the similarity between surrounding pixels and the target pixels by adding the weight into the calculation. With the influence of the weight, this algorithm is able to keep the detailed information in the image.

Its key idea is to get the estimated gray scale of the target pixel from the weighted arithmetic mean of the gray scale of surrounding pixels. As shown in Fig2, for every target pixel x we will have a search window and a neighborhood window.

For every pixel y in the search window, the Euclidean distance $\|V(x)-V(y)\|^2$ is computed. And the weight between them is calculated by

$$w(x,y) = \frac{1}{Z(x)}\exp(-\frac{\|V(x)-V(y)\|^2}{h^2}) \tag{3}$$

$Z(x)$ is used to normalize the weights.

$$Z(x) = \sum_y \exp(-\frac{\|V(x)-V(y)\|^2}{h^2}) \tag{4}$$

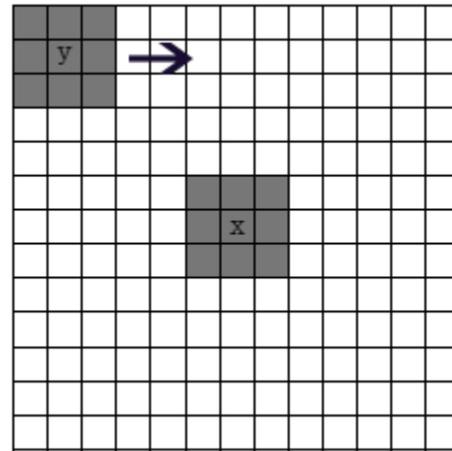

fig2.A direct introduction of the NL-means h here is known as the smooth parameter, which is used to modify the value inside the exp(). When the neighborhood window is big, $\|V(x)-V(y)\|^2$ will be a big number too. Then it will be difficult to see the difference from the w(x,y), because the variance will be very small among pixels with different similarity. And finally the estimation of the gray scale of the target pixel is gained through the following fomula

$$u(x) = \sum_{y \in I} w(x,y)*v(y) \tag{5}$$

## C. Improved core function

Since our image is very small after it has been shrunk into one of four of its original size, it becomes more difficult to keep the edges of the image. So an ideal core function of the NL-means should produce large weight when the distance between two neighborhoods are small and provide very small weight when the distance between two neighborhoods are very large. However, because of the reason mentioned above, each pixel has to go through logarithm at the beginning of the process. As a result, the distance is rather small. Together with the low derivative of the core function of the original NL-means when the variable is smaller than 0, the original method reacts poorly in our algorithm.

$$w(i,j) = \exp(-\frac{D(i,j)}{h^2}) \tag{6}$$

Besides, an appropriate core function is very important when we use our modified NL-means. Since every pixel in the target image will be amplified into four pixel in the algorithm, the sensitivity to the change of the distance of the core function becomes very significant. In a theory[9], the cosine core function was proposed.

$$w(i,j) = \begin{cases} \cos(\frac{\pi D(i,j)}{2h}), & D(i,j) \leq h \\ 0, & D(i,j) > h \end{cases} \tag{7}$$

However, this core function has a problem in its range. In order to perform this algorithm correctly, we have to find an appropriate h. But it is always difficult to find one. Often if the h is too large, the image will blur. Otherwise, if h is too small, many points will be blank.

So in order to solve this problem well, we decide to combine the new core functions and the original core function.

$$w(i,j) = \begin{cases} \exp(\cos(\frac{\pi D(i,j)}{2h_1}))*h_2, & D(i,j) \leq h_1 \\ 0, & D(i,j) > h_1 \end{cases} \tag{8}$$

In this function, even if it is a bit too large, the derivative of the inner function is small but the derivative of the exponent function will make up for this fault. It is used here to modify the value so that we can have a appropriate derivative of the exponential function.

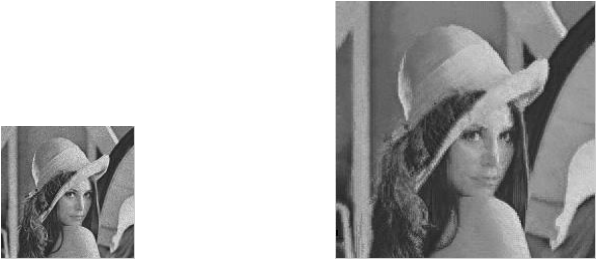

Fig1.The shrunk image with noise

Fig2.The image amplified with modified NL-means with original core function.

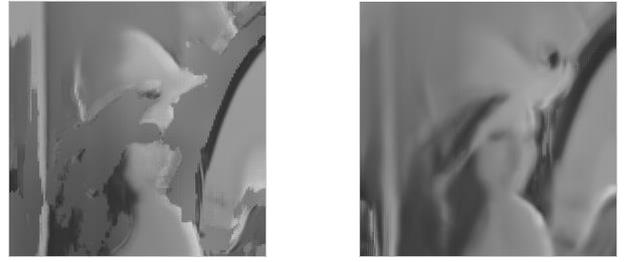

Fig3.What happens if h is too small when using cosine as the new core function.

Fig4.What happens if h is too large when using cosine as the new core function.

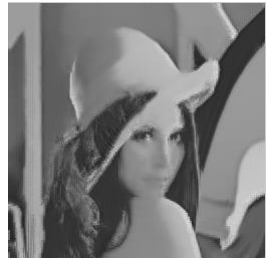

Fig5. The image amplified with modified NL-means with newly proposed core function.

From the above images, we can see that the image with the newly proposed core function has better denoise effect and maintain most of the details. As for the cosine function, it may easily cause the image to either be blur or be blank.

## D. Modified NL-means

In this improved algorithm, we still have a search window and a neighborhood window. However, our search window is in the shrunk image of the original image. Assume that we have a 256*256 original image. Then we will have a searching window consist of surrounding pixels of the pixel which is corresponding to the target pixel after the original image shrunk. The neighborhood window will still be in the original image.

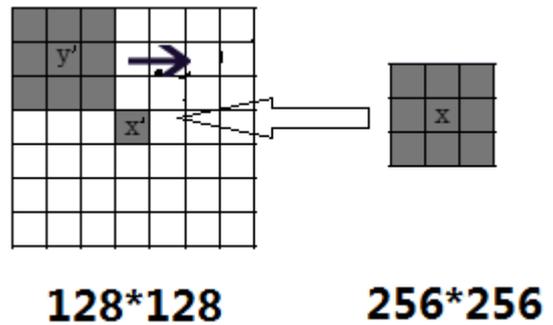

128*128    256*256

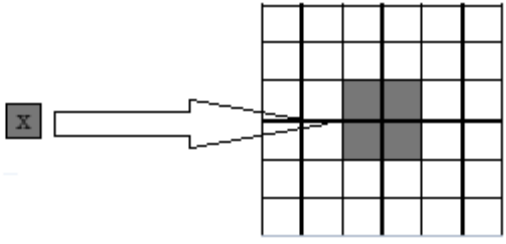

256*256　　　　　512*512

Still, for every pixel y in the search window, the Euclidean distance and the weight will be calculated in the same way as original. Then, as every pixel in 128*128 correspond to four pixels in 256*256, we can gain the four corresponding pixels in 512*512 of the target pixel in 256*256. Denote y', y'', y''', y'''' in original image are the four corresponding pixels of y in the shrunk image, and x', x'', x''', x'''' in the amplified image are the four corresponding pixels of x in the original image. We can modify the NL-means into

$$X'(x) = \sum_{y \in I} w(x,y) * Y'(y) \quad (9)$$

$$X''(x) = \sum_{y \in I} w(x,y) * Y''(y) \quad (10)$$

$$X'''(x) = \sum_{y \in I} w(x,y) * Y'''(y) \quad (11)$$

$$X''''(x) = \sum_{y \in I} w(x,y) * Y''''(y) \quad (12)$$

And the original image will be amplified four times after every pixel has gone through this formula.

*E. Back Propagation Neural Network*

Although modified NL-means can provide a very good result already. We still want to see what if Back Propagation Neural Network is implemented, will it return a better result or not. Let us take a look at a small example to demonstrate the whole process. For example, if we want to reconstruct a image of 256*256, we will first shrink it into a image of 128*128. Then, by applying the modified NL-means mentioned above, we shall get a image of 256*256. By inputting the target pixel in the image of 128*128, and the four pixels in the image corresponding to the target point, and regarding the four pixels in the original image after it is denoised as the output, we can train the neural network. Then we shall input the target point in the original image and the corresponding pixels in the image and the output shall be our result.

## IV. RESULT AND ANALYSIS

To evaluate the effectiveness of the proposed method, experiments are conducted firstly on Lenna with speckle noise then on the real SAR image. Because the SAR image is full of speckle noise, we add speckle noise on the image to simulate SAR image. Then we apply NL-means on the image to see what the best of the method can do. After that, modified NL-means and BP neural network are implemented on the shrunk image separately. Finally, we will propose a combined of modified NL-means and BP neural network to draw a final conclusion.

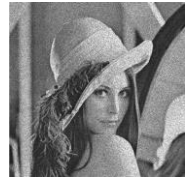 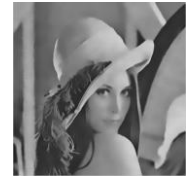

Fig6. Lenna with Speckle noise.　Fig7. Denoise image of fig1.

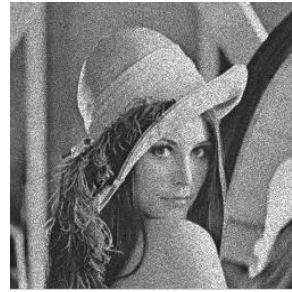 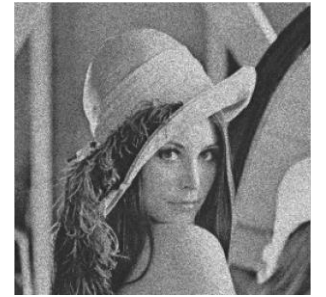

Fig8. modified NL-means result.　Fig9. BP network result.

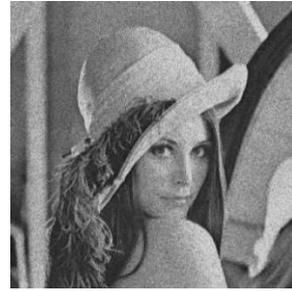

Fig10. combined of NL-means and BP network.

|  | Fig2 | Fig3 | Fig4 | Fig5 |
|---|---|---|---|---|
| PSNR | 27.9027 | 22.5735 | 24.1161 | 25.1140 |
| ENL | 7.1566 | 6.7667 | 7.0037 | 6.9352 |

We can see from the Fig7 that NL-means does a good work in denoising speckle noise. And the modified NL-means also perform well. Fig8 almost keep all the details of the original image although some features have lost due to amplifying the shrunk image. Besides, although we can see from Fig9 BP neural network also performs well without the help of NL-means, combining the these two algorithm still makes an more outstanding result. In Fig10 only 2.79dB was lost after going through shrinking and amplifying again. To make things more specific, we calculate the Equivalent Number of Look(ENL) of the original image and the image mentioned above. The ENL of the original image is 6.7185. We can see that the ENL of our final result is very close to it.

To make it general, let us take a look at the more examples.

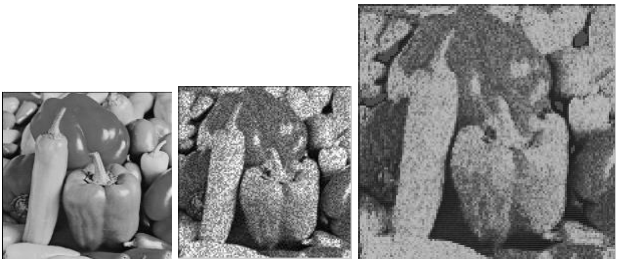

If we apply our algorithm on a real SAR image, we can see that it works well. We can now not only have a higher resolution of the image but also have a clearer view of the scene.

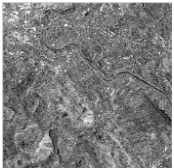
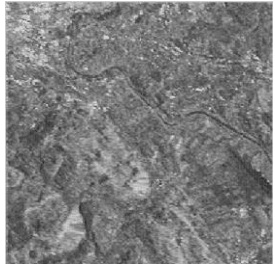

Fig11. SAR image     Fig12.image after this algorithm

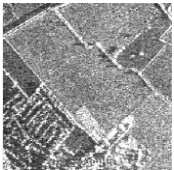
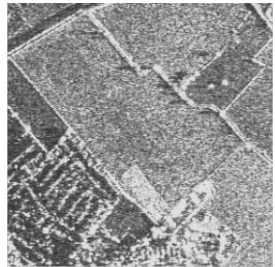

Fig13.SAR Image     Fig14.Image after this algorithm

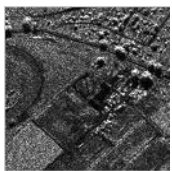
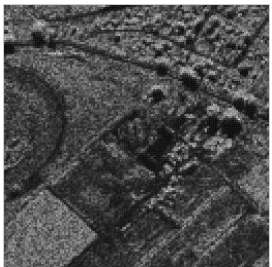

Fig15.SAR Image     Fig16.Image after this algorithm

## V. CONCLUSION

In this paper, a combination of modified NL-means and BP neural network is proposed for super resolution image reconstruction of SAR image. Making use of the redundancy of the image, we can reconstruct the image which maintains lots of details of the original image. With the help of strong nonlinearity mapping ability of BP neural network, we can further improve our result. During the training process, a new core function is used to maintain the edge of shrunk image. The experiment has show that, the modified NL-means not only reconstruct the low-resolution image into high-reconstruction but also denoise the speckle noise in the SAR image and the combination with BP neural network has improved this effect

## *References*

[1] TsaiRY and HuangTS. "Multipleframe image restoration and registration". In Advances in Computer Vision and Image Processing, pages 317(1984)339

[2] Kim, S. P., N. K. Bose, and H. M. Valenzuela. "Recursive reconstruction of high resolution image from noisy undersampled multiframes." IEEE Transactions on Acoustics Speech & Signal Processing38.6(1990):1013-1027.

[3] Wen, Yu Su, and S. P. Kim. "High‐resolution restoration of dynamic image sequences." International Journal of Imaging Systems & Technology 5.4(2005):330-339.

[4] Olshausen, Bruno A. "Emergence of simple-cell receptive field properties by learning a sparse code for natural images." Nature 381.6583 (1996): 607-609.

[5] S. Peleg and M. Irani, "Improving resolution by image registration", CVGIP: Graph. Models Image Processing, 1991, 231-239.

[6] Youla, By D, and H. Webb. "Image restoration by the method of convex projections—part I: theory." IEEE Transactions on Medical Imaging 2010

[7] Stark, H, and P. Oskoui. "High-resolution image recovery from image-plane arrays, using convex projections. " Journal of the Optical Society of America A Optics & Image Science 6.11(1989):1715-1726

[8] Schultz, Richard R., and R. L. Stevenson. "Extraction of high-resolution frames from video sequences." IEEE Transactions on Image Processing A Publication of the IEEE Signal Processing Society 5.6(1996):996-1011.

[9] Johnson, Micah Kimo. "Lighting and optical tools for image forensics." Dartmouth College, 2007.
.